*Research Article*

# Mexican Hat Wavelet Kernel ELM for Multiclass Classification

**Jie Wang, Yi-Fan Song, and Tian-Lei Ma**

*School of Electrical Engineering, Zhengzhou University, Zhengzhou, China*

Correspondence should be addressed to Yi-Fan Song; 574839023@qq.com

Kernel extreme learning machine (KELM) is a novel feedforward neural network, which is widely used in classification problems. To some extent, it solves the existing problems of the invalid nodes and the large computational complexity in ELM. However, the traditional KELM classifier usually has a low test accuracy when it faces multiclass classification problems. In order to solve the above problem, a new classifier, Mexican Hat wavelet KELM classifier, is proposed in this paper. The proposed classifier successfully improves the training accuracy and reduces the training time in the multiclass classification problems. Moreover, the validity of the Mexican Hat wavelet as a kernel function of ELM is rigorously proved. Experimental results on different data sets show that the performance of the proposed classifier is significantly superior to the compared classifiers.

## 1. Introduction

Extreme learning machine, which was proposed by Huang et al. [1] in 2004, is a model of single-hidden layer feedforward neural network. In this model, input weights and hidden layer biases are initialized randomly, and output weights are obtained by using the Moore-Penrose generalized inverse of the hidden layer output matrix. Compared with the conventional BP neural networks, ELM has faster learning speed, higher testing accuracy, and lower computational complexity. Therefore, ELM is widely used in sales forecasting [2], image quality assessment [3], power loss analysis [4], and so on. In 2006, Huang et al. [5] proposed incremental extreme learning machine (I-ELM), which continuously increased the number of hidden layer nodes to improve the training accuracy. Subsequently, Li [6] combined I-ELM with the convex optimization learning method and proposed ECI-ELM in 2014, which reduced the training time of I-ELM. This improvement overcame the weakness of randomly selecting weights in I-ELM and eventually improved the training accuracy. At the same time, Wang and Zhang [7] introduced the Gram-Schmidt orthogonalization method into I-ELM and saved the training time of I-ELM to a large degree. But, in general, I-ELM and its varieties only improve the training accuracy. Their numbers of hidden layer nodes are very likely to exceed the number of samples. Thus, I-ELM greatly improves the training time. In another perspective, in order to achieve a higher training accuracy, Rong et al. [8] used statistical methods to measure the relevance of hidden nodes of ELM and proposed P-ELM in 2008. Then, in 2010, Miche et al. [9] proposed OP-ELM, which is an improvement of P-ELM. In addition, Akusok et al. [10] proposed a high-performance ELM model in 2015, which provides a solid ground for tackling numerous Big Data challenges. However, none of these methods has changed the characteristic of the random selection of input weights. In addition, the linear weighted mapping method in original ELM is not replaced at all.

Therefore, both ELM and its varieties have some inevitable problems. ① Because of the random selection of input weights, some hidden nodes may be given an input weight that is very close to 0, which are commonly called dead nodes. This phenomenon leads to the minimal effect of these nodes and eventually affects the output accuracy. ② With the increment of the number of samples, the hidden nodes number also becomes large. Thus, some high dimensional dot product operations will appear in the training process. Eventually, that will cause the increase of computational complexity and training time. This problem is commonly called dimension explosion. ③ For nonlinear samples, the linear weighted mapping method often has inevitable error, which leads to the reduction of the training accuracy.

In order to solve the above problems, Huang et al. [11] proposed the kernel extreme learning machine (KELM) in 2012, which utilized the kernel function to replace the linear weighted mapping method. Initially, the kernel function they selected is a Gauss function. Although [11] solves the problem of dead nodes and dimension explosion in a sense, the performance of the traditional kernel function for multiclass classification problems is still not very good. From [12, 13], we know that wavelet functions can be used in SVM and ELM, which have a strong fitting capability. Therefore, in this paper, we propose a Mexican Hat wavelet kernel ELM (MHW-KELM) classifier, which effectively solves the problems in the conventional classifier. Compared with the traditional KELM, the MHW-KELM classifier achieves better results on dealing with the multiclass classification problems. Because of that, the new kernel function improves the training accuracy.

The basic principle of ELM and some theorems are shown in Section 2 of this paper. In Section 3, the Mexican Hat wavelet kernel ELM is proposed, and its validity is also proved. Performance evaluation is presented in Section 4. Conclusion is given in Section 5.

## 2. Preliminary Work

*2.1. ELM Model.* Let us suppose that there are arbitrary distinct samples $\{(x_i, t_i) \mid x_i \in R^D, t_i \in R^M, i = 1, 2, \ldots, N\}$. If the number of the hidden nodes is $L$ and the activation function is $g(x)$, then we can randomly select the initial value of the input weights $W$ and the hidden biases $b$. So, the hidden layer output function of ELM can be obtained. It is shown as

$$H = \begin{bmatrix} g(w_1^T x_1 + b_1) & \cdots & g(w_L^T x_1 + b_L) \\ \vdots & \ddots & \vdots \\ g(w_1^T x_N + b_1) & \cdots & g(w_L^T x_N + b_L) \end{bmatrix}, \quad (1)$$

where $w_i \in R^D, b_i \in R, i = 1, 2, \ldots, L$.

If the output weights are $\beta$, according to the proof given by Huang et al. [1], the norm of $\beta$ is smaller, and the generalization performance of ELM is better. Therefore, the output weights $\beta$ can be obtained by finding the least square solution of the problem

$$\text{Minimize:} \quad L_P = \frac{1}{2} \|\beta\|^2 + \frac{C}{2} \sum_{i=1}^{N} \|\xi_i\|^2$$
$$\text{Subject to:} \quad h(x_i)\beta = t_i^T - \xi_i^T, \quad i = 1, 2, \ldots, N, \quad (2)$$

where $h(x_i)$ is the $i$th output vector of hidden layer, $t_i$ is the $i$th label vector, and $\xi_i$ is the error between the $i$th network output vector and the label vector.

According to KKT theory, the above problem can be transformed into a Lagrange function

$$L_D = \frac{1}{2} \|\beta\|^2 + \frac{C}{2} \sum_{i=1}^{N} \|\xi_i\|^2 - \sum_{i=1}^{N} \sum_{j=1}^{M} \alpha_{i,j} \left( h(x_i)\beta_j - t_{i,j} + \xi_{i,j} \right), \quad (3)$$

where each of the Lagrange multipliers $\alpha_i$ corresponds to a sample $x_i$. By calculating the partial derivative of (3), we can get the following set of equations:

$$\frac{\partial L_D}{\partial \beta_j} = 0 \longrightarrow$$
$$\beta_j = \sum_{i=1}^{N} \alpha_i h(x_i)^T = H^T \alpha \quad (4a)$$

$$\frac{\partial L_D}{\partial \xi_i} = 0 \longrightarrow$$
$$\alpha_i = C\xi_i, \quad (4b)$$
$$i = 1, 2, \ldots, N$$

$$\frac{\partial L_D}{\partial \alpha_i} = 0 \longrightarrow$$
$$h(x_i)\beta - t_i^T + \xi_i^T = 0, \quad (4c)$$
$$i = 1, 2, \ldots, N,$$

where $\alpha = [\alpha_1, \ldots, \alpha_N]^T$. And the least square solution of $\beta$ can be obtained by calculating the three equations in (4a), (4b), and (4c). The solution is

$$\hat{\beta} = H^T \left( \frac{I}{C} + HH^T \right)^{-1} T \quad (5)$$

and the output function of ELM is

$$f(x) = h(x) H^T \left( \frac{I}{C} + HH^T \right)^{-1} T. \quad (6)$$

*2.2. Translation-Invariant Kernel Theorem.* Kernel function method is often used in SVM as a method of replacing dot product. According to the Mercer theorem (see [14]), by introducing the kernel function $K(x_i, x_j)$, we can replace the calculation of dot product in ELM. In order to reduce the computational complexity of high dimensional dot product, it is necessary to ensure that $K(x_i, x_j)$ is only a mapping method of the relative position of two input samples (see (7)).

$$K(x_i, x_j) = K(x_i - x_j). \quad (7)$$

The kernel functions which satisfy (7) are called the translation-invariant kernel function. In fact, it is difficult to prove that a translation-invariant kernel function satisfies the Mercer theorem. Fortunately, for the translation-invariant kernel function, the following theorem provides a necessary and sufficient condition to make it become an admissible support vector kernel.

**Theorem 1** (translation-invariant kernel theorem; see [15, 16]). *A translation-invariant kernel $K(x_i, x_j) = K(x_i - x_j)$ is an admissible support vector kernel, if and only if the Fourier transform*

$$F[K](\omega) = (2\pi)^{-D/2} \int_{R^D} \exp(-j\omega x) K(x) dx \quad (8)$$

*is nonnegative.*

The kernel function selection method of ELM is the same as SVM. Therefore, the above theorem can also be used to determine whether a function is an admissible ELM kernel. The commonly used translation-invariant kernel functions are Gauss kernel function and polynomial kernel function. In these two functions, Gauss kernel function is a kind of translation-invariant kernel function. And the expression of the two kernel functions can be given as

$$\text{Gauss: } K(x_i, x_j) = \exp\left(-\frac{\|x_i - x_j\|^2}{2\sigma^2}\right) \quad (9)$$
$$\text{Poly: } K(x_i, x_j) = (1 + x_i x_j)^d.$$

In (9), $\sigma$ is a Gauss core width and $d$ is an adjustable polynomial power exponent.

## 3. Mexican Hat Wavelet Kernel ELM

*3.1. Kernel ELM.* In original ELM model, the linear weighted hidden output function $h(x)$ is usually not satisfied with the mapping method of the nonlinear samples. In order to solve this problem, we can replace $h(x)H^T$ and $HH^T$ in (6) with a kernel function $K(u, v)$. And the result is

$$f(x) = \begin{bmatrix} K(x, x_1) \\ \vdots \\ K(x, x_N) \end{bmatrix}^T \left(\frac{I}{C} + \Omega_{\text{ELM}}\right)^{-1} T, \quad (10)$$

where $\Omega_{\text{ELM}}$ is the kernel function matrix of $X$ (see (11)).

$$\Omega_{\text{ELM}} = [K(x_i, x_j)], \\ i = 1, 2, \ldots, N, \; j = 1, 2, \ldots, N. \quad (11)$$

*3.2. Mexican Hat Wavelet Kernel Function.* In this part, the Mexican Hat wavelet kernel function is proposed. It is also proved that Mexican Hat wavelet function is an admissible ELM kernel.

**Theorem 2** (see [12]). *Let $\psi(x)$ be a mother wavelet. Let $a$ and $c$ denote the dilation and translation, respectively, and $x, a, c \in R$. If $x_i, x_j \in R^D$, then the dot product wavelet kernel is*

$$K(x_i, x_j) = \prod_{d=1}^{D} \psi\left(\frac{x_{di} - c}{a}\right) \psi\left(\frac{x_{dj} - c}{a}\right). \quad (12)$$

If it satisfies the translation-invariant kernel theorem, the following translation-invariant kernel function can be obtained:

$$K(x_i, x_j) = \prod_{d=1}^{D} \psi\left(\frac{x_{di} - x_{dj}}{a}\right). \quad (13)$$

The proof of Theorem 2 is given in [12]; we will not repeat it in this paper. We use Mexican Hat wavelet as the mother wavelet (see (14)). Then, the Mexican Hat wavelet kernel function is derived (see (15)). In this paper, it is also proved that Mexican Hat wavelet satisfies the translation-invariant kernel theorem. In other words, it is also an admissible ELM kernel.

$$\psi(x) = (1 - x^2) \exp\left(-\frac{x^2}{2}\right) \quad (14)$$

$$K(x_i, x_j) = \prod_{d=1}^{D} \left[1 - \left(\frac{x_{di} - x_{dj}}{a}\right)^2\right] \exp\left[-\frac{1}{2}\left(\frac{x_{di} - x_{dj}}{a}\right)^2\right]. \quad (15)$$

**Lemma 3.** *As a kind of translation-invariant kernel function, Mexican Hat wavelet is an admissible ELM kernel.*

*Proof.* Firstly, it should be proved that the Fourier transform of Mexican Hat wavelet is nonnegative (see (16)).

$$F(\omega) = (2\pi)^{D/2} \prod_{d=1}^{D} \int_{-\infty}^{+\infty} \exp(-j\omega_d x_d)\left(1 - \frac{x_d^2}{a^2}\right) \cdot \exp\left(-\frac{x_d^2}{2a^2}\right) dx_d \geq 0. \quad (16)$$

Equation (17) can be decomposed into a set of integral inequalities (see (19)). And the derivation process is

$$F(\omega) = (2\pi)^{D/2} a^D \prod_{d=1}^{D} \int_{-\infty}^{+\infty} \left\{\exp\left[-j\omega a\left(\frac{x_d}{a}\right) - \frac{1}{2}\left(\frac{x_d}{a}\right)^2\right] - \left(\frac{x_d}{a}\right)^2 \exp\left[-j\omega a\left(\frac{x_d}{a}\right) - \frac{1}{2}\left(\frac{x_d}{a}\right)^2\right]\right\} d\left(\frac{x_d}{a}\right)$$
$$= (2\pi)^{D/2} a^D \prod_{d=1}^{D} \exp\left(-\frac{1}{2}a^2\omega_d^2\right) \left\{\int_{-\infty}^{+\infty} \exp\left[-\frac{1}{2}(x_d + ja\omega_d)^2\right] dx_d - \int_{-\infty}^{+\infty} x_d^2 \exp\left[-\frac{1}{2}(x_d + ja\omega_d)^2\right] dx_d\right\}. \quad (17)$$

The integral term in (17) can be written as

$$I = \prod_{d=1}^{D}\left(F_1^{(d)}(\omega) - F_2^{(d)}(\omega)\right), \quad (18)$$

where $I$ is the integral term in (17),

$$F_1^{(d)}(\omega) = \int_{-\infty}^{+\infty} \exp\left[-\frac{1}{2}(x_d + ja\omega_d)^2\right] dx_d, \quad (19)$$

TABLE 1: Basic features of 12 data sets.

| Data set | Training number | Testing number | Attribute | Category |
|---|---|---|---|---|
| Abalone | 2000 | 2177 | 8 | 3 |
| Auto MPG | 200 | 198 | 7 | 5 |
| Bank | 2000 | 2521 | 16 | 2 |
| Car Evaluation | 1000 | 728 | 6 | 4 |
| Wine | 100 | 78 | 13 | 3 |
| Wine Quality | 2000 | 4497 | 11 | 7 |
| Iris | 100 | 50 | 4 | 3 |
| Glass | 100 | 114 | 9 | 2 |
| Image | 100 | 110 | 19 | 7 |
| Yeast | 1000 | 484 | 8 | 4 |
| Zoo | 50 | 51 | 16 | 7 |
| Letter | 2000 | 18000 | 16 | 26 |

$$F_2^{(d)}(\omega) = \int_{-\infty}^{+\infty} x_d^2 \exp\left[-\frac{1}{2}(x_d + ja\omega_d)^2\right] dx_d. \quad (20)$$

According to the translation invariance of the integral, it is easy to get (21) by using the partial integration method. The answer is

$$\begin{aligned} F_1^{(d)}(\omega) &= (2\pi)^{1/2}, \\ F_2^{(d)}(\omega) &= (2\pi)^{1/2}\left(1 - a^2\omega^2\right). \end{aligned} \quad (21)$$

Substituting (21) into (18), we have

$$\prod_{d-1}^{D}\left(F_1^{(d)}(\omega) - F_2^{(d)}(\omega)\right) = (2\pi)^{D/2} a^{2D} \omega^{2D}. \quad (22)$$

Then, substituting (22) into (17), we can obtain the Fourier transform

$$F(\omega) = (2\pi)^D a^{3D} \exp\left(-\frac{a^2}{2}\sum_{d=1}^{D}\omega_d^2\right)\prod_{d=1}^{D}\omega_d^2. \quad (23)$$

From (23), it is known that if $a \geq 0$, $F(\omega) \geq 0$. Therefore, according to the translation-invariant kernel theorem, Mexican Hat wavelet is an admissible ELM kernel. □

3.3. *MHW-KELM Classifier*. We have already proved that Mexican Hat wavelet is an admissible ELM kernel. So, we can substitute (15) into (10) and construct MHW-KELM classifier. For a binary classification problem, the output function of the new classifier is

$$f(x) = \text{sgn}\left\{\begin{bmatrix} K(x,x_1) \\ \vdots \\ K(x,x_N) \end{bmatrix}^T \left(\frac{I}{C} + \Omega_{\text{ELM}}\right)^{-1} T\right\}. \quad (24)$$

Besides, this classifier can also be used for the multiclass classification problems. And the output function is

$$f(x) = \arg\max\left\{\begin{bmatrix} K(x,x_1) \\ \vdots \\ K(x,x_N) \end{bmatrix}^T \left(\frac{I}{C} + \Omega_{\text{ELM}}\right)^{-1} T\right\}. \quad (25)$$

Equation (25) means the classification result is expressed by the index value of the maximum value in output vector. In addition, we can combine the nonnegative constant parameter $a$ of Mexican Hat wavelet and the penalty factor $C$ into an individual and use some evolutionary algorithms such as PSO [17, 18] to find the best values of these parameters. Next, we will analyze the performance of the proposed classifier.

## 4. Performance Evaluation

This section will analyze the performance of MHW-KELM and compare it with the traditional Gauss-KELM, Poly-KELM, original ELM, and BP classifier. All these algorithms run on the R2014a MATLAB software. The operating environment is Core-i7, 2.6 GHz CPU, 8 G RAM. We choose scaled conjugate gradient algorithm to optimize BP neural network, which is faster than normal BP neural network. In order to get excellent performance, the number of hidden nodes of original ELM and BP is selected as 100% and 30% of training samples, respectively. The data sets used in the experiment are from the UCI database [19]. They are Abalone, Auto MPG, Bank, Evaluation, Wine, Wine Quality, Iris, Glass, Image, Yeast, Zoo, and Letter, respectively. The basic features of these 12 data sets are shown in Table 1.

Then, we use the 12 data sets given in Table 1 to test the running time and training accuracy of 5 algorithms. Each data set will be tested by each algorithm 100 times. For each time, the training sample will be randomly selected from the total sample. In order to conduct a rigorous comparison, paired Student's test is performed, which gives the probability that two sets come from distributions with an equal mean. Tables 2–13 record the results of these experiments, and each table corresponds to a data set. All tables have four elements which represent mean accuracy, standard deviation, $p$ value obtained by paired Student's test, and the running time, respectively. For each data set, the data with bold type means this is the best accuracy or the best running time ($p$ value = 1.00), while the data with italic type means there is no statistical difference between this one and the best accuracy or it is very close to the best time ($p$ value $\geq 0.05$).

Table 2: Performance comparison with statistical test on Abalone.

| Data set (training number, category) | | MHW-KELM | Gauss-KELM | Poly-KELM | Original ELM | SCG-BP |
|---|---|---|---|---|---|---|
| Abalone (2000, 3) | Mean | **79.42** | *79.26* | 77.24 | 62.53 | 64.20 |
| | Std. | **±0.36** | *±0.44* | ±0.92 | ±2.89 | ±3.13 |
| | $p$ value | **1.00** | *0.44* | $7.35e-04$ | $2.12e-07$ | $3.47e-07$ |
| | Time | **0.665** | *0.673* | 0.968 | 3.357 | 7.835 |

Table 3: Performance comparison with statistical test on Auto MPG.

| Data set (training number, category) | | MHW-KELM | Gauss-KELM | Poly-KELM | Original ELM | SCG-BP |
|---|---|---|---|---|---|---|
| Auto MPG (200, 5) | Mean | **82.24** | 73.55 | 79.01 | 56.71 | <10 |
| | Std. | **±0.51** | ±1.22 | ±0.90 | ±3.24 | |
| | $p$ value | **1.00** | $1.15e-06$ | $7.35e-04$ | $5.16e-05$ | 0 |
| | Time | 0.070 | 0.071 | 0.075 | **0.058** | 1.235 |

Table 4: Performance comparison with statistical test on Bank.

| Data set (training number, category) | | MHW-KELM | Gauss-KELM | Poly-KELM | Original ELM | SCG-BP |
|---|---|---|---|---|---|---|
| Bank (2000, 2) | Mean | *89.71* | **89.89** | 86.50 | 65.85 | 87.99 |
| | Std. | *±0.43* | **±0.36** | ±0.64 | ±1.43 | ±1.25 |
| | $p$ value | *0.21* | **1.00** | $4.47e-06$ | $1.09e-09$ | 0.03 |
| | Time | *0.657* | **0.652** | 0.8917 | 3.227 | 7.611 |

By drawing the running time in all tables to a line graph, we can get Figure 1. In Figure 1, the horizontal coordinate corresponds to the number of training samples, 50, 100, 200, 1000, and 2000, respectively. Without loss of generality, we can select five data sets, Zoo, Image, Auto MPG, Car Evaluation, and Abalone, as the representations of different numbers of samples. The vertical coordinate shows the mean running time of each data set. Moreover, the running times of MHW-KELM and Gauss-KELM are very close. So, we only draw the running time of MHW-KELM. Four lines are drawn with different styles.

From all tables and Figure 1, it is clear to see that when the training number is larger than 1000, compared to other algorithms, MHW-KELM shows an obvious advantage in running time. For the data sets whose training number is more than 1000, such as Abalone, Bank, Car Evaluation, Wine Quality, Yeast, and Letter, we can obtain that the running time of MHW-KELM and Gauss-KELM is less than that of other algorithms. That means translation-invariant kernel is superior to other kernels. Therefore, it can be concluded that the choice of translation-invariant kernel function can effectively shorten the running time when the training size is large enough.

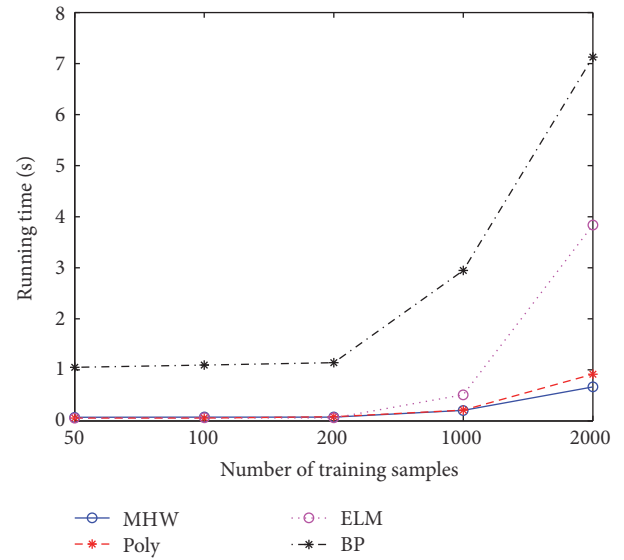

Figure 1: Comparison of running time of 4 algorithms.

From Tables 2–13, it can be obviously seen that the classification performance of MHW-KELM is better than other algorithms when the number of categories is more than 4. The results of paired Student's test show that the performance of MHW-KELM is significantly different ($p$ value $\leq 0.05$) from that of the original ELM and SCG-BP on all data sets, and it is also different from Gauss-KELM and Poly-KELM on Auto MPG, Car Evaluation, Wine Quality, and Image. These four data sets have one thing in common, which is the fact that the category numbers of these data sets are all more than 4.

Table 5: Performance comparison with statistical test on Car Evaluation.

| Data set (training number, category) | | MHW-KELM | Gauss-KELM | Poly-KELM | Original ELM | SCG-BP |
|---|---|---|---|---|---|---|
| Car Evaluation (1000, 4) | Mean | **97.58** | 96.12 | 92.98 | 31.94 | 70.25 |
| | Std. | **±0.38** | ±0.90 | ±1.11 | ±12.36 | ±5.53 |
| | $p$ value | **1.00** | 0.01 | $1.97e-06$ | $8.24e-10$ | $3.68e-08$ |
| | Time | **0.204** | *0.217* | 0.240 | 0.548 | 2.751 |

Table 6: Performance comparison with statistical test on Wine.

| Data set (training number, category) | | MHW-KELM | Gauss-KELM | Poly-KELM | Original ELM | SCG-BP |
|---|---|---|---|---|---|---|
| Wine (100, 3) | Mean | **99.82** | 83.63 | *99.28* | 50.10 | 36.87 |
| | Std. | **±0.09** | ±0.81 | *±0.13* | ±2.93 | ±1.28 |
| | $p$ value | **1.00** | $5.52e-07$ | *0.67* | $4.14e-09$ | $8.05e-10$ |
| | Time | 0.070 | 0.072 | **0.058** | **0.058** | 1.088 |

Table 7: Performance comparison with statistical test on Wine Quality.

| Data set (training number, category) | | MHW-KELM | Gauss-KELM | Poly-KELM | Original ELM | SCG-BP |
|---|---|---|---|---|---|---|
| Wine Quality (2000, 7) | Mean | **54.59** | 49.69 | 52.14 | 45.79 | <10 |
| | Std. | **±0.35** | ±0.52 | ±0.28 | ±0.85 | |
| | $p$ value | **1.00** | $1.04e-06$ | $7.82e-03$ | $3.27e-09$ | 0 |
| | Time | *1.159* | **1.133** | 1.372 | 3.520 | 7.159 |

Table 8: Performance comparison with statistical test on Iris.

| Data set (training number, category) | | MHW-KELM | Gauss-KELM | Poly-KELM | Original ELM | SCG-BP |
|---|---|---|---|---|---|---|
| Iris (100, 3) | Mean | *99.20* | **99.32** | 98.85 | 61.34 | 35.41 |
| | Std. | *±0.16* | **±0.11** | ±0.12 | ±0.78 | ±0.33 |
| | $p$ value | *0.62* | **1.00** | 0.01 | $4.59e-05$ | $6.21e-08$ |
| | Time | 0.071 | 0.075 | 0.062 | **0.055** | 1.290 |

Table 9: Performance comparison with statistical test on Glass.

| Data set (training number, category) | | MHW-KELM | Gauss-KELM | Poly-KELM | Original ELM | SCG-BP |
|---|---|---|---|---|---|---|
| Glass (100, 2) | Mean | 98.11 | *98.47* | **99.41** | 92.83 | 75.76 |
| | Std. | ±0.31 | *±0.42* | **±0.35** | ±1.79 | ±3.63 |
| | $p$ value | 0.02 | *0.06* | **1.00** | $3.50e-05$ | $9.93e-07$ |
| | Time | 0.072 | 0.074 | 0.065 | **0.057** | 1.074 |

Table 10: Performance comparison with statistical test on Image.

| Data set (training number, category) | | MHW-KELM | Gauss-KELM | Poly-KELM | Original ELM | SCG-BP |
|---|---|---|---|---|---|---|
| Image (100, 7) | Mean | **93.87** | 85.12 | 87.58 | 35.56 | 16.13 |
| | Std. | **±1.54** | ±0.78 | ±0.46 | ±1.94 | ±3.25 |
| | $p$ value | **1.00** | $6.64e-06$ | $1.78e-06$ | $8.91e-09$ | $2.45e-11$ |
| | Time | 0.075 | 0.072 | 0.061 | **0.056** | 1.193 |

Table 11: Performance comparison with statistical test on Yeast.

| Data set (training number, category) | | MHW-KELM | Gauss-KELM | Poly-KELM | Original ELM | SCG-BP |
|---|---|---|---|---|---|---|
| Yeast (1000, 4) | Mean | *66.78* | *66.87* | **67.39** | 37.23 | 33.95 |
| | Std. | *±0.72* | *±0.94* | **±0.36** | ±3.71 | ±2.11 |
| | $p$ value | *0.06* | *0.11* | **1.00** | $1.57e-07$ | $7.84e-08$ |
| | Time | **0.193** | 0.201 | 0.235 | 0.457 | 3.005 |

Table 12: Performance comparison with statistical test on Zoo.

| Data set (training number, category) | | MHW-KELM | Gauss-KELM | Poly-KELM | Original ELM | SCG-BP |
|---|---|---|---|---|---|---|
| Zoo (50, 7) | Mean | **99.27** | *99.12* | *99.15* | 92.30 | 35.56 |
| | Std. | **±0.14** | *±0.12* | *±0.13* | ±0.53 | ±1.21 |
| | $p$ value | **1.00** | *0.74* | *0.81* | $1.50e-03$ | $1.58e-04$ |
| | Time | 0.075 | 0.076 | **0.057** | **0.056** | 1.135 |

Table 13: Performance comparison with statistical test on Letter.

| Data set (training number, category) | | MHW-KELM | Gauss-KELM | Poly-KELM | Original ELM | SCG-BP |
|---|---|---|---|---|---|---|
| Letter (2000, 26) | Mean | *72.62* | **86.80** | 68.79 | 15.51 | <10 |
| | Std. | *±13.26* | **±4.32** | ±3.88 | ±5.48 | |
| | $p$ value | *0.11* | **1.00** | 0.01 | $2.43e-03$ | 0 |
| | Time | **1.668** | 1.833 | 2.132 | 4.559 | 7.270 |

Besides, when the category number is less than 4, such as Abalone, Bank, Wine, Iris, Yeast, and Letter, MHW-KELM still has a similar performance to Gauss-KELM or Poly-KELM. Therefore, MHW-KELM is an excellent classifier in multiclass classification problems, which is better than traditional kernel ELM. That means the Mexican Hat wavelet function is a better ELM kernel than the Gaussian function.

## 5. Conclusion

In this paper, we propose a classifier, the Mexican Hat wavelet kernel ELM classifier, which can be applied to the multiclass classification problem. Besides, its validity as an admissible ELM kernel is also proved. This classifier solves the inevitable problems in original ELM by replacing the linear weighted mapping method with Mexican Hat wavelet. The experimental results show that the training time of MHW-KELM classifier is much less than that of original ELM, which solves the problem of the dimension explosion in original ELM. Meanwhile, the training accuracy of this classifier is superior to the traditional Gauss-KELM and original ELM in dealing with the multiclass classification problems.

In future work, in order to reduce the impact of inequality of the training data on the performance, we plan to utilize the boosting weighted ELM proposed by Li et al. [20] to optimize the proposed classifier. In addition, from the experimental results of this paper, it can be seen that a single kernel function cannot meet the requirements of all data sets. So, we are prepared to combine multiple kernel functions to construct mixed kernel ELM, in order to suit different situations.

## Competing Interests

The authors declare that there are no competing interests regarding the publication of this paper.

## Acknowledgments

The authors gratefully acknowledge the support of the following foundations: 973 Project of China (2013CB733605), the National Natural Science Foundation of China (21176073 and 61603343), and the Fundamental Research Funds for the Central Universities.

## References

[1] G. B. Huang, Q. Y. Zhu, and C. K. Siew, "Extreme learning machine: a new learning scheme of feedforward neural networks," in *Proceedings of the International Joint Conference on Neural Networks*, vol. 2, no. 2, pp. 985–990, Budapest, Hungary, 2004.

[2] Z.-L. Sun, T.-M. Choi, K.-F. Au, and Y. Yu, "Sales forecasting using extreme learning machine with applications in fashion retailing," *Decision Support Systems*, vol. 46, no. 1, pp. 411–419, 2008.

[3] S. Suresh, R. Venkatesh Babu, and H. J. Kim, "No-reference image quality assessment using modified extreme learning machine classifier," *Applied Soft Computing Journal*, vol. 9, no. 2, pp. 541–552, 2009.

[4] A. H. Nizar, Z. Y. Dong, and Y. Wang, "Power utility nontechnical loss analysis with extreme learning machine method," *IEEE Transactions on Power Systems*, vol. 23, no. 3, pp. 946–955, 2008.

[5] G.-B. Huang, L. Chen, and C.-K. Siew, "Universal approximation using incremental constructive feedforward networks with random hidden nodes," *IEEE Transactions on Neural Networks*, vol. 17, no. 4, pp. 879–892, 2006.

[6] Y. Li, "Orthogonal incremental extreme learning machine for regression and multiclass classification," *Neural Computing & Applications*, vol. 27, no. 1, pp. 111–120, 2016.

[7] W. Wang and R. Zhang, "Improved convex incremental extreme learning machine based on enhanced random search," in *Unifying Electrical Engineering and Electronics Engineering*, vol. 238 of *Lecture Notes in Electrical Engineering*, pp. 2033–2040, Springer, New York, NY, USA, 2014.

[8] H.-J. Rong, Y.-S. Ong, A.-H. Tan, and Z. Zhu, "A fast pruned-extreme learning machine for classification problem," *Neurocomputing*, vol. 72, no. 1–3, pp. 359–366, 2008.

[9] Y. Miche, A. Sorjamaa, P. Bas, O. Simula, C. Jutten, and A. Lendasse, "OP-ELM: optimally pruned extreme learning machine," *IEEE Transactions on Neural Networks*, vol. 21, no. 1, pp. 158–162, 2010.

[10] A. Akusok, K.-M. Bjork, Y. Miche, and A. Lendasse, "High-performance extreme learning machines: a complete toolbox for big data applications," *IEEE Access*, vol. 3, pp. 1011–1025, 2015.

[11] G.-B. Huang, H. Zhou, X. Ding, and R. Zhang, "Extreme learning machine for regression and multiclass classification," *IEEE Transactions on Systems, Man, and Cybernetics, Part B: Cybernetics*, vol. 42, no. 2, pp. 513–529, 2012.

[12] L. Zhang, W. Zhou, and L. Jiao, "Wavelet support vector machine," *IEEE Transactions on Systems, Man, and Cybernetics, Part B: Cybernetics*, vol. 34, no. 1, pp. 34–39, 2004.

[13] W. Qin, S. Yuantong, K. Yu, W. Qiang, and S. Lin, "Parsimonious wavelet kernel extreme learning machine," *Journal of Engineering Science and Technology Review*, vol. 8, no. 5, pp. 219–226, 2015.

[14] J. Mercer, "Functions of positive and negative type, and their connection with the theory of integral equations," *Philosophical Transactions of the Royal Society A: Mathematical, Physical and Engineering Sciences*, vol. 209, no. 441-458, pp. 415–446, 1909.

[15] A. J. Smola, B. Schölkopf, and K.-R. Müller, "The connection between regularization operators and support vector kernels," *Neural Networks*, vol. 11, no. 4, pp. 637–649, 1998.

[16] C. J. C. Burges, "Geometry and invariance in kernel based methods," in *Advances in Kernel Methods*, MIT Press, 1999.

[17] J. Kennedy and R. Eberhart, "Particle swarm optimization," in *Proceedings of the IEEE International Conference on Neural Networks*, pp. 1942–1948, December 1995.

[18] R. Eberhart and J. Kennedy, "A new optimizer using particle swarm theory," in *Proceedings of the 6th International Symposium on Micro Machine and Human Science (MHS '95)*, pp. 39–43, Nagoya, Japan, October 1995.

[19] M. Lichman, *UCI Machine Learning Repository*, University of California, School of Information and Computer Science, Irvine, Calif, USA, 2013, http://archive.ics.uci.edu/ml.

[20] K. Li, X. Kong, Z. Lu, L. Wenyin, and J. Yin, "Boosting weighted ELM for imbalanced learning," *Neurocomputing*, vol. 128, pp. 15–21, 2014.

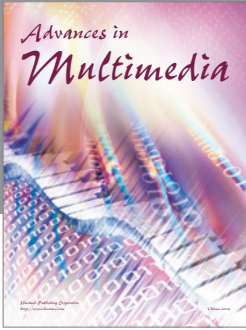
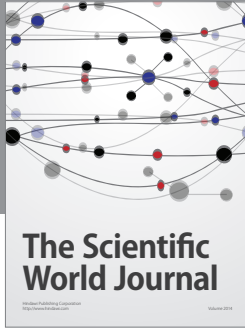
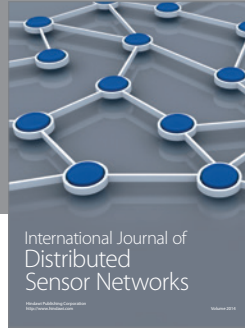
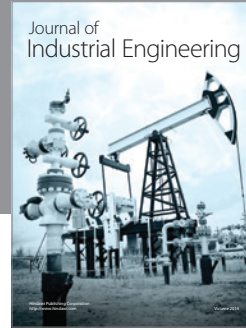
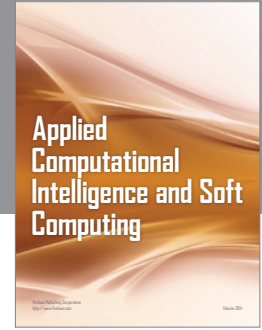
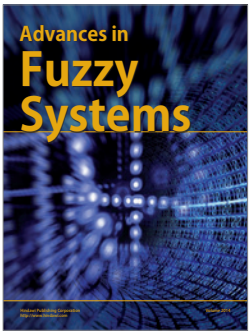
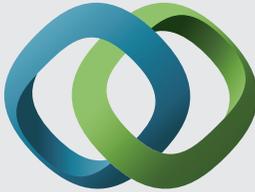
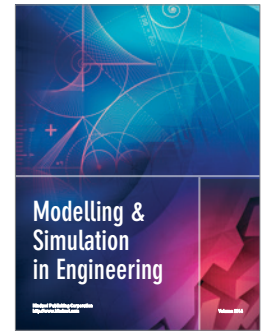
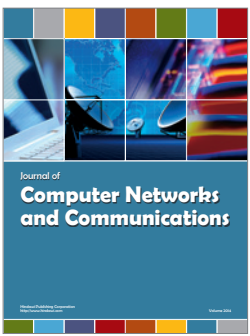
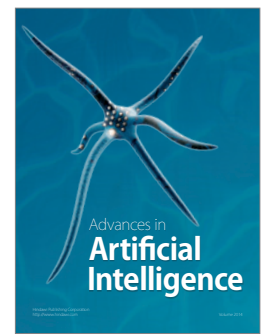
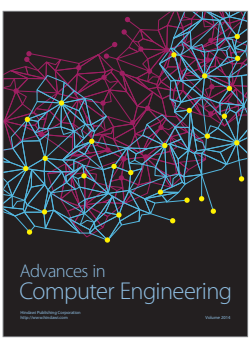
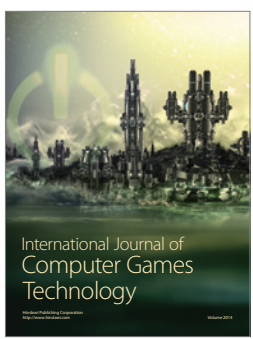
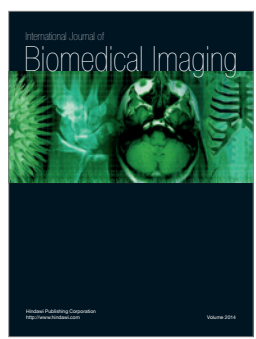
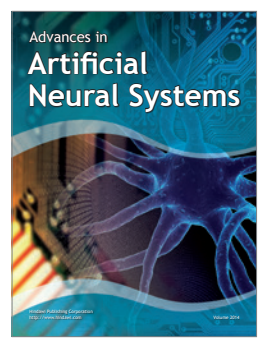
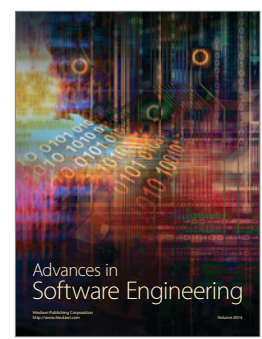
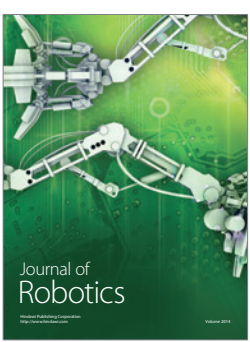
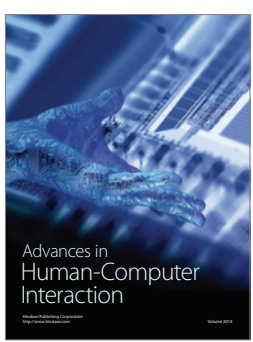
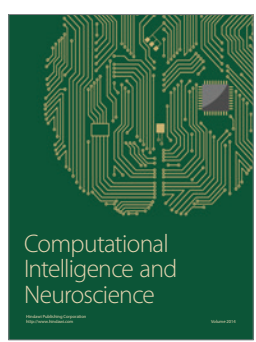
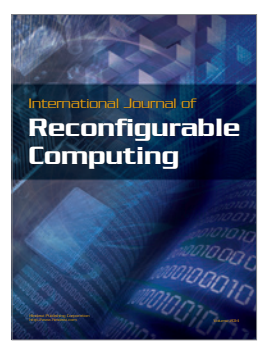
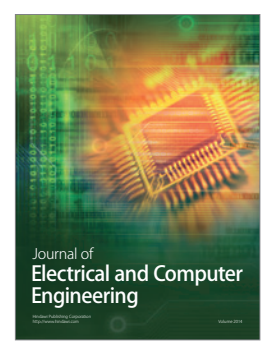